\documentclass[runningheads]{llncs}

\usepackage[T1]{fontenc}

\usepackage{graphicx}
\usepackage{amssymb}
\usepackage{amsmath}
\usepackage{bbm}
\usepackage{tikz}
\usepackage{booktabs}
\usepackage{listings}
\usepackage{hyperref}
\usetikzlibrary{shapes.geometric, arrows, positioning}

\usepackage{color}

\newcommand{\G}{\mathcal{G}}
\newcommand{\R}{\mathcal{R}}
\newcommand{\V}{\mathcal{V}}
\newcommand{\X}{\mathcal{X}}
\newcommand{\F}{\mathcal{F}}

\newcommand{\spo}{\langle s, p, o\rangle}
\newcommand{\query}{\langle s, p, ?\rangle}

\newcommand{\benchmark}{\textsc{GRainsaCK}}
\newcommand{\kelpiepp}{\textsc{Kelpie{\scriptsize ++}}}

\begin{document}

\title{\benchmark{}: a Comprehensive Software Library for Benchmarking Explanations of Link Prediction Tasks on Knowledge Graphs}

\titlerunning{A Library for Benchmarking Explanations of Link Predictions}

\author{
    Roberto Barile\inst{1}\orcidID{0009-0007-3058-8692} \and 
    Claudia d'Amato\inst{1,2}\orcidID{0000-0002-3385-987X} \and 
    Nicola Fanizzi\inst{1,2}\orcidID{0000-0001-5319-7933}
}

\institute{
    Dipartimento di Informatica -- University of Bari Aldo Moro, Italy\\ 
    \email{r.barile17@phd.uniba.it}, \email{claudia.damato@uniba.it}, \email{nicola.fanizzi@uniba.it} \and 
    CILA -- University of Bari Aldo Moro, Italy
}

\maketitle

\begin{abstract}
Since Knowledge Graphs are often incomplete, link prediction methods are adopted for predicting missing facts. 
Scalable embedding based solutions are mostly adopted for this purpose, however, they lack comprehensibility, which may be crucial in several domains.
Explanation methods tackle this issue by identifying supporting knowledge explaining the predicted facts.
Regretfully, evaluating/comparing quantitatively the resulting explanations is challenging as there is no standard evaluation protocol and overall benchmarking resource.
We fill this important gap by proposing \benchmark{}, a reusable software resource that fully streamlines all the tasks involved in benchmarking explanations, i.e., from model training to evaluation of explanations along the same evaluation protocol.
Moreover, \benchmark{} furthers modularity/extensibility by implementing the main components as functions that can be easily replaced.
Finally, fostering its reuse, we provide extensive documentation including a tutorial.
\end{abstract}

\section{Introduction}
\label{sect:introduction}
\emph{Knowledge Graphs} (KGs) are formal machine-processable representations of knowledge~\cite{hogan_knowledgegraphs_2022} that conform to graph-based data models consisting of entities (nodes) and binary relations (edges).
KGs often deliver, in addition to facts, intensional knowledge, which enables sound reasoning and is typically represented via ontologies.
Despite their proven utility in academic and business initiatives~\cite{pellissiertanonYAGOReasonableKnowledge2020,dongBuildingBroadKnowledge2019}, KGs are often noisy and/or incomplete because the activities characterizing their life-cycle are often semi-automatic, incremental, and distributed~\cite{hogan_knowledgegraphs_2022}.
\emph{Link Prediction} (LP) methods aim at completing KGs by predicting missing facts and mostly ground on \emph{Knowledge Graph Embedding} (KGE) models that lead to competitive accuracy and scalability~\cite{rossiKnowledgeGraphEmbedding2021}.
KGE models are representation learning solutions that encode the elements of a KG as low-dimensional vectors, preserving their structural properties, that can be leveraged for tackling complex downstream tasks, such as LP, using efficient linear algebra operations.
Despite such advantages, these models lack comprehensibility, i.e., are not traceable in terms of operations on symbolic/explicit knowledge such as KG facts and ontological axioms.
This problem hampers the use of LP via KGE models particularly in fields where it is paramount that stakeholders can comprehend predictions before relying on them for making decisions with critical consequences.
For example, LP may be used for predicting the side effects of a drug~\cite{novavcek2020predicting}, but it is crucial that stakeholders comprehend the predictions before relying on them for making decisions about funding of research on the drug.

LP eXplanation (LP-X) methods~\cite{schramm_comprehensibleartificial_2023} address this issue.
Specifically, a post-hoc (after the prediction) LP-X method works with a generic LP solution and explains a prediction by computing pieces of knowledge (e.g., sets of facts) that are associated (via the LP method) to the prediction.
As also emerged during the  AAAI 2025 Presidential Panel~\footnote{https://aaai.org/about-aaai/presidential-panel-on-the-future-of-ai-research/}, \textit{``principled empirical evaluation} [is] \textit{more important than ever''} also for explanations.
For this purpose, we target benchmarking LP-X methods via a standard framework for quantitative evaluation and comparison of different LP-X methods, thus allowing to prove the validity and generality of LP-X proposals.
In fact, conducting such evaluations is challenging due to the lack of a standard evaluation protocol, gold standards, and resources for benchmarking LP-X methods.
For filling this gap, we aim at engineering a comprehensive software library for benchmarking LP-X methods.
Indeed, existing benchmarking libraries~\cite{le2023benchmarking} mostly target images and tabular data and as such do not fit LP-X.
Moreover, the few existing approaches~\cite{rossi_explaininglink_2022,dixit} for evaluating explanations computed via LP-X methods (e.g., fidelity, length) do not adopt the same evaluation protocol and although provided with prototype implementations are not intended as reusable or extensible software libraries.

With this work, we pursue the following goals.
First, we aim to formalize an automated workflow that specifies the tasks involved in benchmarking LP-X methods.
Indeed, state-of-the-art approaches lack a clear workflow or, when defined, tasks have to be initiated manually after verifying that the dependencies are met.
Second, we address the formalization of a modular design and implementation of the different components of the proposed workflow, thus supporting the maintainability and extensibility of the evaluation process, e.g., the integration of novel LP-X methods.
Existing methods present duplicated implementations that make extension and maintenance difficult. 
Moreover, they often do not reuse state-of-the-art open source libraries, e.g., implementing LP, rather they implement such components from scratch.
Hence, they cannot benefit from the integration of ongoing novel approaches.
The expected scientific impact is the standardization of the LP-X benchmarking resources and practices, including evaluation protocols and metrics.

To further these goals, we propose \benchmark{}: an open source software library for benchmarking LP-X methods that is applicable to RDF/RDFS/OWL KGs and most of the existing KGE and LP-X methods.
\benchmark{} is based on the theoretical formulation of LP-DIXIT~\cite{dixit}, to the best of our knowledge, the sole existing method for measuring the quality of an explanation (of a LP task) that is user guided yet fully algorithmic and works for explanations coming from a generic LP-X method.
LP-DIXIT grounds on the \textit{Forward Simulatability Variation} (FSV) measure~\cite{nautaAnecdotalEvidenceQuantitative2023} which is based on a cognitive theory stating that predictions are understandable when simulatable.
The FSV often underpins user studies, but LP-DIXIT bypasses the need for experts by employing \emph{Large Language Models} (LLMs) in order to mimic actual users.
\benchmark{} rationalizes the benchmarking of explanations by formalizing the workflow underlying two types of experiments involving LP-DIXIT.
Firstly, it targets experiments for validating the hypothesis that LLMs can mimic users in evaluating explanations, by measuring the agreement of LP-DIXIT with expert-curated ground-truth datasets.
Secondly, it targets the experiments for comparing different LP-X methods via LP-DIXIT.
\benchmark{} is developed in Python and adopts a modular architecture where the components are implemented as functions that can be easily replaced or extended.
In order to maximize software reuse, \benchmark{} also relies on and extends PyKEEN~\cite{ali2021pykeen}, a state-of-the-art library for KGE learning and LP.
In order to facilitate the (re)use of \benchmark{}, it is accompanied by extensive documentation including a tutorial for installation, usage (as an API or via CLI commands), and extension.

The rest of the paper is organized as follows.
In \S~\ref{sect:related_works}, we review state-of-the-art methods and resources for computing and evaluating explanations.
In \S~\ref{sect:proposal}, we detail \benchmark{}, while in \S~\ref{sect:evaluation}, we illustrate its evaluation.
In \S~\ref{sect:conclusions}, we summarize the achievements and suggest future research.


\section{Related Works}
\label{sect:related_works}
This section analyzes the state-of-the-art resources for benchmarking explanations and the LP-X approaches, along with the methods for evaluating their explanations.

Several libraries for benchmarking explanations~\cite{le2023benchmarking} have been developed but none of them target LP-X.
\textsc{Quantus}~\cite{quantus} supports the largest variety of explanation types (not including LP-X) and also targets diverse dimensions of explanation quality~\cite{nautaAnecdotalEvidenceQuantitative2023}.
In contrast, most of the existing tool kits (see the survey in~\cite{le2023benchmarking}), such as \textsc{GraphXAI}~\cite{agarwal2023evaluating} and \textsc{OpenXAI}~\cite{agarwal2022openxai}, mainly target limited types of explanations (not including LP-X methods for KGs), e.g., feature importance, and measure the quality of explanations basically as the fidelity to the prediction.
None of them targets the utility of the explanation from the user's perspective with the FSV. 

Another direction is to provide datasets with ground-truth explanations to be compared with the computed ones.
FR200K~\cite{halliwellUserScoredEvaluation2021}, FRUNI and FTREE~\cite{martin_fruniftree_2023}, include hand-crafted rules that reflect domain knowledge and explain a fact by identifying the facts that underpin the rule generating that fact.
In FR200K each explanation is also rated by users in terms of (subjective) intuitiveness.
Hence, FR200K enables user guided evaluation; however, its construction process hardly generalizes to large scale due to the required manual intervention.

As for LP-X solutions, that we specifically target as they are generic with respect to the LP method, they explain a prediction by computing pieces of knowledge (e.g., sets of facts) that are associated to the prediction.
The first proposals explain a prediction by returning exactly one fact (within the KG), as in the case of \textsc{DP}~\cite{zhang_datapoisoning_2019}, applying perturbations, or \textsc{Criage}~\cite{pezeshkpour_investigatingrobustness_2019} computing (approximate) influence functions.
The latter can be restricted to a limited set of facts and to specific classes of KGE models.
More recent methods explain a prediction by returning a set of facts.
\textsc{Kelpie}~\cite{rossi_explaininglink_2022} and \kelpiepp{}~\cite{barileExplanationLinkPredictions2024} employ a \emph{post-training} process.
\textsc{KE-X}~\cite{zhao_kexsubgraph_2023} is based on information gain and \textsc{KGExplainer}~\cite{ma_kgexplainerexploring_2024} adopts greedy search and perturbations.
Notably, \textsc{GEnI}~\cite{amador-dominguez_geniframework_2023} returns explanations also including high-level schema axioms based on numerical criteria on (specific classes of) KGEs.
Conversely, the method introduced in~\cite{betz_adversarialexplanations_2022} grounds on abduction via learned rules.
The resulting explanations are mainly evaluated by \emph{re-training} the KGE model, i.e., by comparing the LP performance of the original model with that of a model trained on a modified KG where the facts in the explanations have been added, removed, or isolated.
This approach is fully algorithmic, but it is not user guided and adopts different fusion strategies and metrics depending on the LP-X method.
In contrast, LP-DIXIT, which is our goal, is user guided, fully algorithmic and works for explanations coming from a generic LP-X method.

\textsc{CrossE}~\cite{zhang_interactionembeddings_2019} and \textsc{SemanticCrossE}~\cite{damato_approachbased_2021} explain a prediction by identifying a path between the entities in the prediction.
They rely on similarity measures and evaluate explanations as the number of similar paths connecting similar entities.
Other methods return explanations other than sets of facts or paths.
For example, in~\cite{krishnan_modelagnosticmethod_2023} logical rules are mined to explain a set of predictions and are evaluated in terms of classification performance on the explained predictions and synthetic negative (false) facts.
With \textsc{FeaBI}~\cite{ismaeil_feabifeature_2023}, interpretable vectors are extracted from KGEs via feature selection and are compared to those learned with an interpretable LP method.
The evaluation measures the influence of the LP explanations on the solution of related tasks, without considering the user's perspective.
These evaluation protocols do not allow comparing the explanations coming from the different approaches.

A complementary direction is represented by interpretable LP methods, which are LP methods with a more understandable functioning.
A comparison of different interpretable methods would mean to compare their functioning and is beyond our purpose.


\section{The Proposed Library \benchmark{}}\label{sect:proposal}
We propose \benchmark{}: an open source software library for benchmarking LP-X methods.
It grounds on the theoretical formulation of the LP-DIXIT method for evaluating explanations of LP tasks. 
\benchmark{} is developed in Python, adopts the functional paradigm, and reuses existing software components.
\benchmark{} also gathers and make easily available explanation ground-truth datasets.
\benchmark{} streamlines benchmarking LP-X methods by formalizing the experiment workflow.
It also features several alternative implementations for each task in the workflow, with the possibility of extending them as needed.
Specifically, \benchmark{} supports two types of experiments:
\begin{description}
    \item[validation] measure the agreement of LP-DIXIT with ground-truth datasets that provide expert-curated explanations and their evaluations,
    \item[comparison] compare different LP-X methods via LP-DIXIT.
\end{description}

In \S~\ref{subsect:evaluation_methods}, we illustrate the KGs, ground-truth datasets and KGE models included in \benchmark{}.
In \S~\ref{subsect:evaluation_methods}, we summarize LP-DIXIT and provide the description of the implemented framework and related metrics.
In \S~\ref{subsect:workflow}, we illustrate the formalized workflow.
In \S~\ref{subsect:explanation_methods}, we detail the implemented state-of-the-art LP-X methods.

\subsection{The Supported Datasets and Models}\label{subsect:data}
\benchmark{} includes a curated collection of KGs to be used in the \textbf{comparison} experiments and explanation ground-truth datasets to be used in the \textbf{validation} experiments.

A KG is a graph-based data structure $\G(\V, \R)$, where $\V$ is a set of nodes representing entities, and $\R$ is a set of predicates or edge types, representing binary relations between entities.
A KG can be considered as a collection of triple statements $\spo  \in \V \times \R \times \V$, with a \textit{subject} $s$, a \textit{predicate} $p$ and an \textit{object} $o$, where $s, o \in \V$ and $p \in \R$.
As for the KGs, \benchmark{} relies on PyKEEN for KG loading and thus supports any KG provided by or loadable within PyKEEN.
Additionally, for KGs endowed with schema level information, the types/classes of entities retrieved and inferred from the OWL ontologies are also considered, as for the case of the KGs DB50K, DB100K, and YAGO4-20 that have been enriched with the DBpedia Ontology and YAGO 4 ontology, respectively~\cite{barileExplanationLinkPredictions2024}.

A ground-truth dataset consists of explanations, each one associated to a prediction and a corresponding quality measure of the explanation, as provided by a verifier/user.
A ground-truth dataset can be formalized as a KG and two vectors: the vector of explanations $\vec{x} = (x_1, x_2, \ldots, x_n)$ where each one is associated to a triple representing a prediction and often consists of a set of triples in the KG; and the corresponding vector of explanation quality measures $\vec{f} = (f_1, f_2, \ldots, f_n)$, where each $f_i \in \{\,-1, 0, 1\,\}$ and $1$ indicates a very useful explanation, $0$ a neutral explanation, and $-1$ a useless or misleading explanation.
\benchmark{} currently includes the ground-truth datasets FR200K~\cite{halliwellUserScoredEvaluation2021}, FRUNI~\cite{martin_fruniftree_2023}, and FTREE~\cite{martin_fruniftree_2023}.
FR200K is a sub-graph of DBpedia focused on the French royal families.
FRUNI and FTREE are synthetic KGs modeling relationships among university students and family trees, respectively.
They include hand-crafted rules (implemented either as logical clauses or as Python code) that reflect domain knowledge and explain a predicted triple by identifying the triples in the KG that underpin the rule generating that triple.
For FRUNI and FTREE, explanations are not assessed by users, rather they are assumed to be valuable as derived from rules that capture the domain knowledge accurately.
In contrast, each explanation in FR200K is rated by users in terms of (subjective) intuitiveness that is expressed in the interval $[0, 1]$.
We performed a quantile-based discretization of the ratings into the categorical values $\{\,-1, 0, 1\,\}$ where, for this dataset, $1$ indicates a very intuitive explanation, $0$ a somehow intuitive explanation, and $-1$ a not intuitive explanation.
FR200K contains $2125$ entities, $6$ relations, and $12357$ triples.
In contrast, FRUNI and FTREE are synthetic KGs and their generation (via the tool kit released in~\cite{martin_fruniftree_2023}) is configurable, hence their statistics depend on the specific generation configuration.
We make the enriched KGs and FR200K available on Figshare.

As for the LP methods, they compute the following functions: $LP\colon \V \times \R \to \V$ for completing the given query as an incomplete triple such as $\query$;
and $rank\colon \V \times \R \times \V \to \mathbbm{N}$ for evaluating the LP performance and compute the \emph{rank} of the given triple.
\benchmark{} relies on PyKEEN for LP methods based on KGE models (requiring KGs formatted as specified by PyKEEN, i.e., as three sets of triples, namely training, validation, and test set), and as such supports any LP method implemented in PyKEEN, e.g., \textsc{TransE}~\cite{bordes_translatingembeddings_2013} and \textsc{ComplEx}~\cite{trouillon_complexembeddings_2016}, which is itself easily extensible.


\subsection{Implementing LP-DIXIT Evaluation Protocol and Metrics}
\label{subsect:evaluation_methods}

\benchmark{} employs LP-DIXIT~\cite{dixit} as the theoretical method for measuring the quality of explanations of LP tasks.
It measures the \emph{Forward Simulatability Variation} (FSV) induced by an explanation for a prediction (made by a LP method), i.e., the variation between the simulatability (or predictability) of a prediction without and with an explanation.
A prediction is simulatable (with an explanation) if a (human) verifier can correctly simulate the prediction, i.e., can hypothesize the output of the LP method given the same input provided to the (LP) method (and the explanation).
In the following, we formalize the FSV.

Let $\X$ be the set of the explanations (e.g., sets of triples) for a predicted triple $\spo$, and $S\colon (\V \times \R) \times (\X \cup \{\,\emptyset\,\}) \to \V$ be the function denoting the verifier that simulates the prediction.
It returns, for a query (and an explanation), an entity (being the filler for the query) estimating the one to be returned by the LP method.
Let $I\colon (\V \times \R) \times \V \to \{\,0, 1\,\}$ denote the correctness of a simulation for a prediction, i.e., whether the output of the simulation matches the one of the LP method.
$I$ can be defined via the indicator function:
$$
\forall q := \query \in (\V \times \R), \forall e \in \V\colon I(q, e) = \mathbbm{1}_{LP(q)}(e)
$$
As such, $I$ returns $1$ if the simulation is correct, $0$ otherwise.
Finally, let $F\colon (\V \times \R) \times (\X \cup \{\,\emptyset\,\}) \to \{\,-1, 0, 1\,\}$, be the function measuring the FSV as the difference between the correctness of the simulation with the explanation and the correctness of the simulation without the explanation\footnote{In this case the empty explanation $\emptyset$ is considered.}, formally:
\begin{align*}
\forall q := \query \in (\V \times \R), \forall x \in \X\colon \\
F(q, x) = I(q, S(q, x)) - I(q, S(q, \emptyset)) = \mathbbm{1}_{LP(q)}(S(q, x)) - \mathbbm{1}_{LP(q)}(S(q, \emptyset)) 
\end{align*}
The values returned by $F$ are to be interpreted as follows:
\begin{itemize}
\item $1$, the explanation is \emph{beneficial} for the verifier: the simulation with the explanation $x$ is correct, instead, the one without $x$ is incorrect (i.e., $I(q, S(q, \emptyset)) = 0, I(q, S(q, x)) = 1$).
\item $0$, the explanation is \emph{neutral} for the verifier: either both simulations are correct (i.e., $I(q, S(q, \emptyset)) = 1, I(q, S(q, x)) = 1$) or both are incorrect (i.e., $I(q, S(q, \emptyset)) = 0, I(q, S(q, x)) = 0$).
\item $-1$, the explanation is \emph{harmful} for the verifier: the simulation with the explanation $x$ is incorrect, instead, the one without $x$ is correct (i.e., $I(q, S(q, \emptyset)) = 1, I(q, S(q, x)) = 0$).
\end{itemize}

LP-DIXIT employs a LLM as the verifier for computing $S$.
Specifically, four alternative prompting methods are considered: zero-shot (task input only) without output constraints, few-shot (task input and examples) without output constraint, zero-shot with output constraints, and few-shot with output constraints.
LP-DIXIT defines a general prompt template that is instantiated according to the selected prompting method and the explanation to be evaluated, as illustrated in Fig.~\ref{lst:prompt-template}.
\begin{figure}[tb]
\centering
\scriptsize
\begin{lstlisting}
You are a helpful, respectful and honest assistant.
Your response should be crisp, short and not repetitive.
Discard any preamble, explanation, greeting, or final consideration.

A triple is a statement <subject, predicate, object>.
The subject and the object are entities, and the predicate is a relation from the subject to the object.
Perform a Link Prediction task, given a query as an incomplete triple <subject, predicate, ?>, predict the missing object that completes the triple making it a true statement.
Strict requirement: output solely the name of a single object entity, discard any explanation or other text. 
Correct format: Elizabeth_of_Bohemia
Incorrect format: The object entity is Elizabeth_of_Bohemia.

{examples}

({subject}, {predicate}, ?)
{explanation}

{output_constraint}
\end{lstlisting}
\caption{Prompt template used by LP-DIXIT with prompt sections separated by blank lines and variable parts enclosed in curly braces.}
\label{lst:prompt-template}
\end{figure}
Its first section represents the instruction, i.e., the description of the LP task to be performed/simulated.
It specifies the syntax of a triple and defines LP as returning the (name of) the entity that fills a query, i.e., a triple with an unknown object.
The second section of the prompt includes output formatting instructions, along with an example, directing the LLM to return only the entity name.
Otherwise, the LLM may generate additional text, whilst in the formalization of the FSV it is assumed that the verifier returns solely an entity.
The output constraint mitigates the issue of LLMs that may generate answers that are not entities included in the KG, it consists of a subset of the entities in the KG and an instruction forcing the LLM to pick its answer from such a subset.
LP-DIXIT further contextualizes the LLM on the LP task to be simulated by adopting few-shot prompting, i.e., by including in the prompt a set of examples of solved LP queries.

\benchmark{} implements LP-DIXIT via \textsc{huggingface}, a state-of-the-art library for natural language processing, also supporting LLM prompting.
\benchmark{} also employs \textsc{unsloth} that makes LLM prompting more efficient with no modifications to the implementation.
As such, \benchmark{} supports any LLM coming from \textsc{unsloth}, e.g., the Llama3 herd of models~\cite{grattafiori2024llama}.
Given a set of predictions with associated explanations, \benchmark{} transforms each item into a prompt by filling the template.
\benchmark{} verbalizes explanations, i.e., encodes them as text to be included in the prompt, either verbatim, preserving the original entity and predicate labels, or via a custom verbalization function specified alongside the LP-X method.
Moreover, \benchmark{} processes the prompts in batches, where the batch size is a hyperparameter, in order to maximize parallelization.

\benchmark{} implements different metrics corresponding to the \textbf{validation} and \textbf{comparison} experiments, respectively.
All metrics are based on the output of the function $F$, that is computed on a vector of explanations $\vec{x} = (x_1, x_2, \ldots, x_n)$ and that returns a vector $\hat{\vec{f}} = (\hat{f}_1, \hat{f}_2, \ldots, \hat{f}_n)$, where each $\hat{f}_i \in \{\,-1, 0, 1\,\}$ is the FSV of the explanation $x_i$.
For the case of \textbf{validation} experiments, \benchmark{} computes the classification report.
Specifically, it evaluates the agreement of LP-DIXIT with ground-truth datasets by comparing the estimated vector of FSV values $\hat{\vec{f}}$ against the ground-truth $\vec{f}$.
This is done by computing standard classification metrics over the labels $\{\,-1, 0, 1\,\}$, e.g., per-class precision, recall, and F$_\beta$-score.
As for the \textbf{comparison} experiments, \benchmark{} computes the average FSV and the FSV distribution.
The average FSV is the average $\overline{f}$ of each $\hat{\vec{f}}$ resulting from evaluating explanations (computing $F$) for each LP-X method.
The average FSV ranges in $[-1, 1]$, where values close to $1$ indicate that explanations are mostly \textit{beneficial} for the verifier, values around $0$ suggest that explanations are mostly \textit{neutral} for the verifier, and values near $-1$ reflect that explanations are mostly \textit{harmful} for the verifier.
However, the average FSV does not distinguish between a set of \textit{neutral} explanations (e.g., $(0, 0, 0, 0)$) and a set containing an equal number of \textit{harmful} and \textit{beneficial} explanations (e.g., $(-1, -1, 1, 1)$), both yielding an average FSV of $0$.
As such, \benchmark{} also supports the distribution of FSV within $\hat{\vec{f}}$, i.e., the proportion of $1$, $0$, and $-1$ values.
This provides additional insight and resolves the ambiguity of the average, but fails to summarize the vector in a single scalar value.

\begin{table}[!ht]
\scriptsize
\centering
\label{tab:tasks}
\caption{Definition of the tasks in \benchmark{}}
\begin{tabular}{rl}
\toprule
               & TuneTask                                                                                                                                              \\
\midrule
\bf Parameters & kg\_name, kge\_name                                                                                                                                   \\
\bf Run        & \texttt{tune}(kg, kge\_name)                                                                                                                          \\
\bf Output     & hp\_config.\{kg\_name\}\_\{kge\_name\}                                                                                                                \\
\bf Requires   & kg                                                                                                                                                    \\
\midrule
               & TrainTask                                                                                                                                             \\
\midrule
\bf Parameters & kg\_name, kge\_name                                                                                                                                   \\
\bf Run        & \texttt{train}(kg, kge\_name, hp\_config.\{kg\_name\}\_\{kge\_name\})                                                                                 \\
\bf Output     & kge.\{kg\_name\}\_\{kge\_name\}                                                                                                                       \\
\bf Requires   & TuneTask(kg\_name, kge\_name)                                                                                                                         \\
\midrule
               & RankTask                                                                                                                                              \\
\midrule
\bf Parameters & kg\_name, kge\_name                                                                                                                                   \\
\bf Run        & \texttt{rank}(kg, kge.\{kg\_name\}\_\{kge\_name\})                                                                                                    \\
\bf Output     & ranked.\{kg\_name\}\_\{kge\_name\}                                                                                                                    \\
\bf Requires   & TrainTask(kg\_name, kge\_name)                                                                                                                        \\
\midrule
               & SelectPredictionsTask                                                                                                                                 \\
\midrule
\bf Parameters & kg\_name, kge\_name\                                                                                                                                  \\
\bf Run        & \texttt{select\_predictions}(ranked.\{kg\_name\}\_\{kge\_name\})                                                                                      \\
\bf Output     & predictions.\{kg\_name\}\_\{kge\_name\}                                                                                                               \\
\bf Requires   & RankTask(kg\_name, kge\_name)                                                                                                                         \\
\midrule
               & ExplainTask                                                                                                                                           \\
\midrule
\bf Parameters & kg\_name, kge\_name, lpx\_config                                                                                                                      \\
\bf Run        & \multicolumn{1}{p{10cm}}{\texttt{explain}(predictions.\{kg\_name\}\_\{kge\_name\}, kg, kge.\{kg\_name\}\_\{kge\_name\}, lpx\_config)}                 \\
\bf Output     & explanations.\{kg\_name\}\_\{kge\_name\}\_\{lpx\_config\}                                                                                             \\
\bf Requires   & SelectPredictionsTask(kg\_name, kge\_name)                                                                                                            \\
\midrule
               & EvaluateTask                                                                                                                                          \\
\midrule
\bf Parameters & kg\_name, kge\_name, lpx\_config, eval\_config                                                                                                        \\
\bf Run        & \multicolumn{1}{p{10cm}}{\texttt{evaluate}(explanations.\{kg\_name\}\_\{kge\_name\}\_\{lpx\_config\}, predictions.\{kg\_name\}\_\{kge\_name\}, eval\_config)} \\
\bf Output     & scores.\{kg\_name\}\_\{kge\_name\}\_\{lpx\_config\}\_\{eval\_config\}                                                                                 \\
\bf Requires   & ExplainTask(kg\_name, kge\_name, lpx\_config)                                                                                                         \\
\midrule
               & MetricsTask                                                                                                                                           \\
\midrule
\bf Parameters & kg\_name, kge\_name, lpx\_config, eval\_config, metric\_names                                                                                         \\
\bf Run        & \multicolumn{1}{p{10cm}}{\texttt{metrics}(metric\_names, scores.\{kg\_name\}\_\{kge\_name\}\_\{lpx\_config\}\_\{eval\_config\})}                      \\
\bf Output     & metrics.\{kg\_name\}\_\{kge\_name\}\_\{lpx\_config\}\_\{eval\_config\}\_\{metric\_names\}                                                             \\
\bf Requires   & EvaluateTask(kg\_name, kge\_name, lpx\_config, eval\_config)                                                                                          \\
\bottomrule
\end{tabular}
\end{table}

\subsection{The Benchmarking Operations and Workflow}\label{subsect:workflow}
\benchmark{} formalizes an end-to-end, reproducible and fully automated workflow for benchmarking LP-X methods via LP-DIXIT.
Given an experimental setup, \benchmark{} executes all necessary steps, from data loading to metric computation, with a single command either via CLI or API. 
Specifically, an experimental setup is a list of tests each specifying: a KG name, a KGE model name, an explanation configuration, an evaluation configuration, a metric name.
The structure of the workflow is the same for the two types of experiments, but is instantiated differently, specifically with a different metric and in the \textbf{validation} experiments with the LP-X method (in the explanation configuration) implicitly set to \textit{ground-truth}, meaning that explanations and evaluations are retrieved from a ground-truth dataset.

To achieve this, \benchmark{} defines a set of operations to be performed in benchmarking LP-X methods, each of them formalized and implemented as a function.
In the following, we describe and formalize the signature of each function (operation), i.e., the formal interface that defines the input/output types.
Let $\mathsf{KG}$ be the space of all KGs, $\mathsf{Config}$ be the space of key-value configurations (e.g., for specifying hyperparameters), $\mathsf{KGE}$ be the space of KGE models, $\mathsf{RankedKG}$ be the space of KGs where each triple is associated with its rank, $\mathsf{Explanation}$ be the space of explanations (e.g., sets of triples).
At first, $\texttt{tune}$ selects for the KGE model specified in the given config the best hyperparameter config based on the performance on the given KG, formally:
$$
\texttt{tune}\colon \mathsf{KG} \times \mathsf{Config} \to \mathsf{Config}.
$$
Next, $\texttt{train}$ trains on the given KG the KGE model specified in the given config along with the hyperparameters, formally:
$$
\texttt{train}\colon \mathsf{KG} \times \mathsf{Config} \to \mathsf{KGE}.
$$
Moreover, \texttt{rank} computes the rank of each triple in the given KG via the given KGE model, formally:
$$
\texttt{rank}\colon \mathsf{KG} \times \mathsf{KGE} \to \mathsf{RankedKG}.
$$
\benchmark{} reuses from PyKEEN the implementation of $\texttt{tune}$ along with the possible hyperparameter values, $\texttt{train}$, and $\texttt{rank}$.
Additionally, $\texttt{select\_predictions}$ selects the top ranked triples from the given ranked triples, formally:
$$
\texttt{select\_predictions}\colon \mathsf{RankedKG} \to \mathsf{KG}.
$$
Next, $\texttt{explain}$ computes the explanations for the given KG (predictions) using the statements in the other given KG and based on the given KGE model (used for making the predictions) and explanation config, formally:
$$
\texttt{explain}\colon \mathsf{KG} \times \mathsf{KG} \times \mathsf{KGE} \times \mathsf{Config} \to 2^\mathsf{Explanation}.
$$
In addition, $\texttt{evaluate}$ evaluates the given explanations for the given KG (predictions) according to the given config, formally:
$$
\texttt{evaluate}\colon 2^\mathsf{Explanation} \times \mathsf{KG} \times \mathsf{Config} \to \mathbbm{R}^n.
$$
where $n$ is the number of explanations and predictions.
Finally, $\texttt{metrics}$ aggregates the result of the evaluation of multiple explanations according to the given config, formally:
$$
\texttt{metrics}\colon \mathbbm{R}^n \times \mathsf{Config} \to \mathbbm{R}.
$$

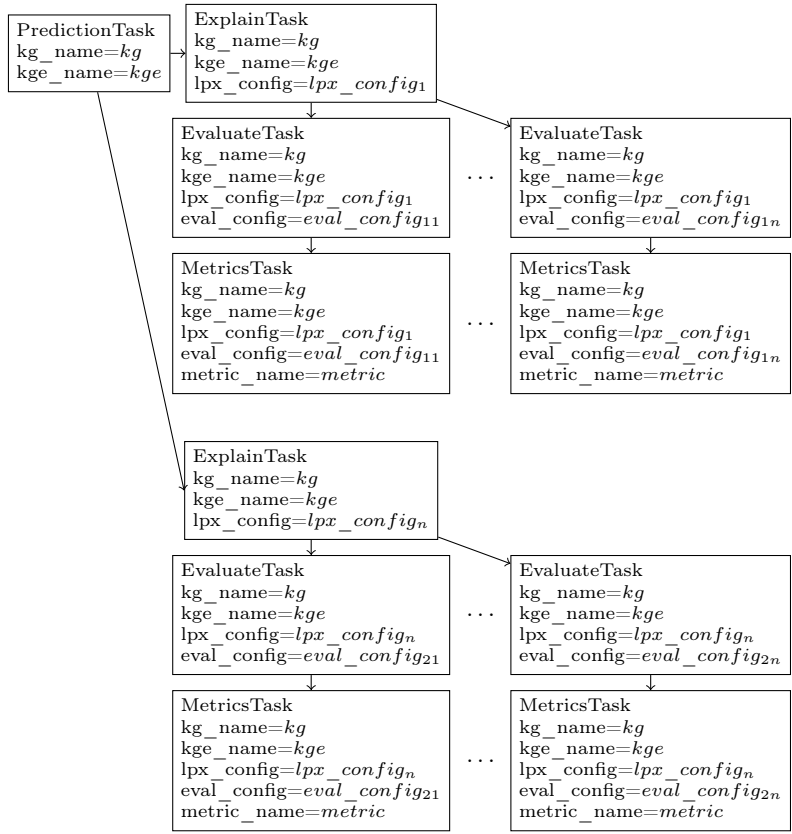
\begin{figure}[!ht]
    \centering
    \begin{tikzpicture}[
      task/.style={draw, rectangle, align=left},
      ellipsis/.style={font=\footnotesize},
      every node/.style={font=\scriptsize}
    ]
        \node[task] (gp1) {PredictionTask\\kg\_name=$kg$\\kge\_name=$kge$};
        
        \node[task, right=2mm of gp1] (explain11) {ExplainTask\\kg\_name=$kg$\\kge\_name=$kge$\\lpx\_config=$lpx\_config_1$};
        \node[task, below=45mm of explain11] (explain12) {ExplainTask\\kg\_name=$kg$\\kge\_name=$kge$\\lpx\_config=$lpx\_config_n$};
        
        \node[task, below=2mm of explain11] (fsv111) {EvaluateTask\\kg\_name=$kg$\\kge\_name=$kge$\\lpx\_config=$lpx\_config_1$\\eval\_config=$eval\_config_{11}$};
        \node[ellipsis, right=1mm of fsv111] {$\dots$};
        \node[task, right=8mm of fsv111] (fsv112) {EvaluateTask\\kg\_name=$kg$\\kge\_name=$kge$\\lpx\_config=$lpx\_config_1$\\eval\_config=$eval\_config_{1n}$};

        \node[task, below=2mm of fsv111] (metrics111) {MetricsTask\\kg\_name=$kg$\\kge\_name=$kge$\\lpx\_config=$lpx\_config_1$\\eval\_config=$eval\_config_{11}$\\metric\_name=$metric$};
        \node[ellipsis, right=1mm of metrics111] {$\dots$};
        \node[task, right=8mm of metrics111] (metrics112) {MetricsTask\\kg\_name=$kg$\\kge\_name=$kge$\\lpx\_config=$lpx\_config_1$\\eval\_config=$eval\_config_{1n}$\\metric\_name=$metric$};
        
        \node[task, below=2mm of explain12] (fsv121) {EvaluateTask\\kg\_name=$kg$\\kge\_name=$kge$\\lpx\_config=$lpx\_config_n$\\eval\_config=$eval\_config_{21}$};
        \node[ellipsis, right=1mm of fsv121] {$\dots$};
        \node[task, right=8mm of fsv121] (fsv122) {EvaluateTask\\kg\_name=$kg$\\kge\_name=$kge$\\lpx\_config=$lpx\_config_n$\\eval\_config=$eval\_config_{2n}$};

        \node[task, below=2mm of fsv121] (metrics121) {MetricsTask\\kg\_name=$kg$\\kge\_name=$kge$\\lpx\_config=$lpx\_config_n$\\eval\_config=$eval\_config_{21}$\\metric\_name=$metric$};
        \node[ellipsis, right=1mm of metrics121] {$\dots$};
        \node[task, right=8mm of metrics121] (metrics122) {MetricsTask\\kg\_name=$kg$\\kge\_name=$kge$\\lpx\_config=$lpx\_config_n$\\eval\_config=$eval\_config_{2n}$\\metric\_name=$metric$};
        
        \draw[->] (gp1) -- (explain11);
        \draw[->] (gp1) -- (explain12.west);
        \draw[->] (explain11) -- (fsv111);
        \draw[->] (explain11) -- (fsv112);
        \draw[->] (explain12) -- (fsv121);
        \draw[->] (explain12) -- (fsv122);
        \draw[->] (fsv111) -- (metrics111);
        \draw[->] (fsv112) -- (metrics112);
        \draw[->] (fsv121) -- (metrics121);
        \draw[->] (fsv122) -- (metrics122);        
    \end{tikzpicture}

    \caption{Representation of the DAG obtained by instantiating the workflow in GRainsaCK with an experimental setup}
    \label{fig:workflow_dag}
\end{figure}

Based on the declared functions, \benchmark{} defines the workflow template as a set of tasks, i.e., units of work that specify the input parameters, i.e., the experimental setup fields on which it depends, the function to be executed, the output, and the tasks it requires to be completed prior to its execution.
Tab.~\ref{tab:tasks} illustrates the tasks in \benchmark{}.
\benchmark{} defines the tasks using \textsc{luigi}, a state-of-the-art workflow orchestration system that allows tasks to be specified entirely within Python.
Given the experimental setup, \benchmark{} instantiates from the workflow template a \emph{Directed Acyclic Graph} (DAG), where the nodes represent the tasks and the edges represent the dependencies among them.
Specifically, a directed edge from node $a$ to node $b$ indicates that task $a$ must be completed before task $b$.
Fig.~\ref{fig:workflow_dag} illustrates the DAG obtained by instantiating the workflow from a given experimental setup.
The DAG is executed by resolving dependencies in topological order. 
As such, \benchmark{} supports intermediate result caching, i.e., reusing of task outputs that are already available.
It also supports deduplication of shared tasks and parallel execution of independent tasks.
For example, if multiple LP-X methods are specified for the same KG and KGE model, the training, and prediction are performed only once and (re)used across all explanation tasks, and the explanation tasks are performed in parallel.

\subsection{The Implemented Explanation Methods}\label{subsect:explanation_methods}
\benchmark{} implements the LP-X methods \textsc{Criage}, \textsc{DP}, \textsc{Kelpie} and \kelpiepp{} as these are the most general with respect to the different LP solutions.
\benchmark{} re-frames their different formalizations to a unified abstraction based on combinatorial optimization, i.e., finding the best solution in a finite set of possible solutions.
As such, an LP-X method consists of different components, each formalized as a function.

In the following we formalize the signature of each function for a KG $\G(\V, \R)$ and describe, as an example, how \textsc{Kelpie} implements them.
Preliminarily, let
$$
\X := \bigcup_{i=1}^{k} \binom{\G}{i}
$$
be the set of all the explanations, where $\binom{G}{i}$ is the set of combinations of size $i \leq k, i,k \in \mathbbm{N}$ and without repetitions of the triples in $\G$.
First, let 
$$\texttt{possible\_explanations}\colon (\V \times \R \times \V) \to 2^\X$$
be the function selecting for the given prediction a (sub)set of possible explanations, where $2^\X$ is the set of all the subsets of $\X$.
\textsc{Kelpie} and \kelpiepp{}, given a prediction $\spo$, performs the following steps: 1)~extract from $\G$ the sub-graph $\G^s$ of all the triples featuring $s$ as subject or as object, 2)~filter this sub-graph to obtain $\F^s$ by selecting the top-$k$ most fitting triples based on paths in the KG, 3)~combines the triples in $\F^s$.
\kelpiepp{} also applies graph summarization in order to decrease the number of triples in $\F^s$ and consequently the number of possible explanations.
Second, let
$$
\texttt{best\_explanation}\colon (\V \times \R \times \V) \times 2^\X \to \X
$$
be the function selecting, for the given prediction, the best explanation in a finite (sub)set of possible explanations based on an objective function.
Finally, let 
$$\texttt{relevance}\colon (\V \times \R \times \V) \times \X \to \mathbbm{R}$$
be the objective function.
The relevance of a possible explanation can be either: 1)~necessary, i.e., with respect to the prediction, 2)~sufficient, i.e., with respect to a set of triples $\{\,\langle c, p, o\rangle \mid c \in C \subset \V\,\}$, where $C$, in the current implementation, is selected as the set of entities for which $LP(\langle c, p, ?\rangle) \neq o$.
\textsc{Kelpie} implements both relevance functions based on partial re-training of KGE models.

\benchmark{} also include multiple baseline LP-X methods that can be employed as references to be compared with more complex LP-X methods, such as those based on combinatorial optimization.
They select a random set of $k$ triples from those involving either 1)~the subject of the prediction, 2)~the predicate of the prediction, or 3)~the object of the prediction.


\section{Evaluation}\label{sect:evaluation}
\subsection{Usage and Extension Proof of Concept}\label{sect:usage}
We present a proof of concept that demonstrates how \benchmark{} simplifies the benchmarking of LP-X methods by showcasing an example of its usage.
Additionally, we illustrate an example regarding the extension of the library with new LP-X methods.

The first step is to easily install \benchmark{} via the Python package manager \texttt{pip} by running the CLI command:
\begin{verbatim}
pip install grainsack
\end{verbatim}
After installing the package, the next step is to download the necessary data to the working directory.
The entire setup for the \textbf{validation} and \textbf{comparison} experiments can be defined in simple \texttt{CSV} files.
Tab.~\ref{tab:example_validation_experimental_setup} and Tab.~\ref{tab:example_comparison_experimental_setup} show examples for \textbf{validation} and \textbf{comparison} experiments, respectively.
These example configurations involve various combinations of KGs and KGE models.
For the \textbf{comparison} experiments, they also include different LP-X methods and evaluation protocols.
All of these can be executed together within a single benchmarking workflow.
Once the configuration file is specified, the entire workflow can be executed with a single command.
Specifically, for the \textbf{validation} experiments the CLI command is: 
\begin{verbatim}
grainsack validation
\end{verbatim}
Similarly, for the \textbf{comparison} experiments the CLI command is:
\begin{verbatim}
grainsack comparison
\end{verbatim}

\begin{table}
    \scriptsize
    \centering
    \caption{Example Experimental Setup for the validation experiments}
    \label{tab:example_validation_experimental_setup}
    \begin{tabular}{lllll}
    \toprule
    \bf kg\_name & \bf kge\_name & \bf eval\_config            \\
    \midrule
    FR200K       & \sc ComplEx   & \{ prompting=zero\_shot, llm=Llama3.1 \} \\
    FRUNI        & \sc ComplEx   & \{ prompting=zero\_shot, llm=Llama3.1 \} \\
    FTREE        & \sc TransE    & \{ prompting=zero\_shot, llm=Llama3.1 \} \\
    \bottomrule
    \end{tabular}
\end{table}

\begin{table}
    \scriptsize
    \centering
    \caption{Example Experimental Setup for the comparison experiments}
    \label{tab:example_comparison_experimental_setup}
    \begin{tabular}{lllll}
    \toprule
    \bf kg\_name & \bf kge\_name & \bf lpx\_config             & \bf eval\_config            \\
    \midrule
    DB100K       & \sc ComplEx   & \{ method=\textsc{Kelpie} \} & \{ prompting=zero\_shot, llm=Llama3.1 \} \\
    DB100K       & \sc ComplEx   & \{ method=\textsc{Criage} \} & \{ prompting=zero\_shot, llm=Llama3.1 \} \\
    DB100K       & \sc TransE    & \{ method=\textsc{Kelpie} \} & \{ prompting=zero\_shot, llm=Llama3.1 \} \\
    YAGO4-20     & \sc ComplEx   & \{ method=\textsc{Kelpie} \} & \{ prompting=zero\_shot, llm=Llama3.1 \} \\
    YAGO4-20     & \sc ComplEx   & \{ method=\textsc{Criage} \} & \{ prompting=zero\_shot, llm=Llama3.1 \} \\
    YAGO4-20     & \sc TransE    & \{ method=\textsc{Kelpie} \} & \{ prompting=zero\_shot, llm=Llama3.1 \} \\
    \bottomrule
    \end{tabular}
\end{table}

This launches a complete end-to-end workflow that encompasses all required phases, from KG loading to metrics computation and result aggregation.
All the metrics are aggregated and saved in a \texttt{JSON} file in the working directory.
Intermediate results are automatically stored for reuse, e.g., in subsequent executions of workflows with different experimental setups, for example.
Additionally, the program supports the deduplication of shared tasks and the parallel execution of independent tasks.
For example, in the \textbf{comparison} setup, multiple LP-X methods are applied to predictions made via \textsc{ComplEx} on DB100K.
Therefore, training and prediction are executed only once and can be reused across all explanation tasks.
Moreover, the different LP-X methods are executed in parallel.

\benchmark{} can also be executed as a Python API.
This enables the extension of \benchmark{}, e.g., implementing custom LP-X methods as functions. 
An example is reported in Fig.~\ref{fig:extension}.
In the example, two custom LP-X methods (\texttt{method\_1}, \texttt{method\_2}) are declared as functions.
Additionally, a function factory (\texttt{explain\_factory}) is used to return the appropriate function based on the method name specified in the \texttt{CSV} file of the experimental setup.

\begin{figure}[!ht]
\begin{lstlisting}[language=Python,basicstyle=\tt\scriptsize]
def method_1(prediction):
    pass

def method_2(prediction):
    pass

def explain_factory(method):
    if method == "method_1":
        return method_1
    else:
        return method_2

luigi.build([Comparison("explain_factory")])
\end{lstlisting}

\label{fig:extension}
\caption{Example of extension of \benchmark{} with new LP-X methods}
\end{figure}

\subsection{Static Code Analysis}
We illustrate the outcomes of the static code analysis of \benchmark{}, using \textsc{pylint}, a state-of-the-art static code analyzer that identifies potential errors and suggests refactoring, assigning a numerical score between $0$ (poor) and $10$ (excellent). 
The analysis yielded a score of $9.49$, did not report any error, and highlighted limited refactoring suggestions, summarized in the following.

A remark concerns the high branching of the factory function, i.e., the function that compose the components of LP-X methods in a $\texttt{explain}$ function, based on the explanation configuration.
This is because the implemented LP-X methods have diverse configurable components.
Related to this, several remarks also point to functions with more than $5$ arguments, as for the case of the $\texttt{explain}$ function which is highly modular, i.e., the components of LP-X methods can be replaced with different implementations, thus requiring each component to be an argument. 
Another common remark concerns the number of local variables within functions exceeding $15$.
This is because we store intermediate results in separate variables for improving readability. 
\benchmark{} follows the functional paradigm except for its integration with \textsc{PyKEEN}, which follows the object-oriented paradigm.
In this respect, some remarks relate to class hierarchies being too deep; they manly stem from extending \textsc{PyKEEN}, which itself defines a deep class hierarchy.


\section{Conclusions}\label{sect:conclusions}
We introduced \benchmark{}, an open source software library for benchmarking LP-X methods.
It employs LP-DIXIT~\cite{dixit} as the method for measuring the quality of explanations of LP tasks since, to the best of our knowledge, it is the sole existing one that is user guided yet fully algorithmic and works for explanations coming from a generic LP-X method.
\benchmark{} specifically formalizes the workflow of two types of experiments: \textbf{validation} experiments, for measuring the agreement of LP-DIXIT with ground-truth datasets; \textbf{comparison} experiments, for comparing different LP-X methods via LP-DIXIT.
\benchmark{} is developed in Python and adopts a modular architecture following the functional paradigm, implementing components as functions that can be easily replaced.
We presented a proof of concept demonstrating how \benchmark{} simplifies benchmarking.
We additionally performed static code analysis via \textsc{pylint}.
For the future, we aim at extending \benchmark{} with additional LP-X methods and releasing the obtained benchmark/evaluation.

\paragraph*{Resource Availability Statement:}
Source code and documentation for \benchmark{} are available from GitHub~\footnote{https://github.com/rbarile17/grainsack} and Zenodo~\footnote{https://doi.org/10.5281/zenodo.15411798} (the most up-to-date information is on GitHub).
Datasets FR200K, DB50K, DB100, and YAGO4-20 are available on Figshare~\footnote{https://doi.org/10.6084/m9.figshare.29066612.v1}.

\begin{credits}
    \subsubsection{\ackname}
    This work was partially supported by project \emph{FAIR - Future AI Research} (PE00000013), spoke 6 - Symbiotic AI (\url{https://future-ai-research.it/}) and by PRIN project \emph{HypeKG - Hybrid Prediction and Explanation with Knowledge Graphs} (Prot. 2022Y34XNM, CUP H53D23003700006), both under the PNRR MUR program funded by the European Union - NextGenerationEU.
\end{credits}

\bibliographystyle{splncs04}
\bibliography{references}

\end{document}